\documentclass[runningheads]{llncs}
\usepackage[T1]{fontenc}
\usepackage{graphicx}
\usepackage{hyperref}
\usepackage{amssymb}
\usepackage{booktabs}
\usepackage{placeins}
\usepackage{float}
\begin{document}

\title{Existential Crisis: A Social Robot's Reason for Being}

\author{Dora Medgyesy \and Joella Galas \and Julian van Pol \and
Rustam Eynaliyev \and Thijs Vollebregt }

\authorrunning{Medgyesy et al.}

\institute{Vrije Universiteit Amsterdam}
\maketitle

\begin{abstract}
As Robots become ever more important in our daily lives there's growing need for understanding how they're perceived by people. This study aims to investigate how the user perception of robots is influenced by displays of personality. Using LLMs and speech to text technology, we designed a within-subject study to compare two conditions: a personality-driven robot and a purely task-oriented, personality-neutral robot. Twelve participants, recruited from Socially Intelligent Robotics course at Vrije Universiteit Amsterdam, interacted with a robot Nao tasked with asking them a set of medical questions under both conditions. After completing both interactions, the participants completed a user experience questionnaire measuring their emotional states and robot perception using standardized questionnaires from the SRI and Psychology literature.
\end{abstract}

\section{Introduction}

Recent Large Language Models (LLM), (GPT4 \cite{GPT4}, Llama \cite{Llama2024},) have enabled unlocked new conversational capabilities, reaching wide-ranging applications in fields as diverse as coding, healthcare, and education \cite{Leng2024}. These new capabilities present an opportunity to accelerate the field of Socially Intelligent Robotics, enabling robots to more intelligently respond to user interactions.\\

There are existing occurrences of robots being applied in medical settings. For example, Pepper robot \cite{Pepper} is capable of talking to patients and enabling them to provide feedback on how they feel through a touch screen, going through a questionnaire. Unfortunately, due to lacking turn-taking capabilities, it is currently incapable of going through a conversation. Despite the growing significance of SRIs, there remains an important gap in understanding user perception of robot behavior \cite{Mileounis2015}. This gap is the subject of study in the field of Human-Robot Interaction (HRI).\\

Personality in humans can be described as a set of long-term affective traits. \cite{picard}. It influences our emotions and through them how we think, act, and react to our surroundings. \cite{Minsky2006}. \\

In robots, personality is typically conveyed via gestures, tone of voice, and choice of language. HRI studies the quality and effectiveness of this interaction. Since non-verbal communication accounts for most human interaction \cite{Mehrabian1967}, it is conceivable that a robot’s personality, as displayed through its tone, behavior, and responsiveness, can significantly influence user perceptions, satisfaction, and overall experience. \cite{Mehrabian1967}
\\\\
This study aims to investigate the impact of personality on HRI in a medical setting. Past studies have developed standardized measures to quantify and compare the robot perception in social settings \cite{Bartneck2009}, \cite{Carpinella2017}, \cite{Heerink2010}. These metrics provide quantified user experience in terms of anthropomorphism, likeability, and perceived intelligence. These dimensions are crucial for evaluating how the introduction of personality in robots affects user experience. By incorporating these measures into the study design we aim to arrive at a more thorough understanding of user experience.\\

With the recent advances in large foundation models trained on Internet-scale data opening up new opportunities for enhancing social interaction far beyond the traditional narrow, pre-programmed expert agents we can now build robots capable of more dynamic, contextually adaptive behaviors, enabling new personality-driven interactions.
\\\\
Our study leverages the latest LLMs and standardized questionnaires to investigate the impact of robot personality, and particularly humor on user experience. Our research question is: \textit{To what extent does adding personality to a questionnaire robot increase user experience?}. Early studies explored how various traits of a robot influence human perception. These traits range from its visual form \cite{Carpinella2017}, to its personality (proactive vs re-active) to the details of its behavior such as the impact of pauses in its nonverbal interaction \cite{Admoni2014}. With this in mind, our hypothesis is that \textit{Users will react positively to a robot exhibiting personality through incorporating contextual and responsive jokes into the reaction compared to a more formal, task-oriented robot}.

\section{Methodology}

\subsection{Participants}
The participants included in this study consist of twelve students enrolled in the course 'Socially Intelligent Robotics' at the \textit{Vrije Universiteit Amsterdam}. The students were selected by the course coordinators. This study did not have any inclusion or exclusion criteria for the participants. All participants who partook in the experiment managed to fully finish it. All participants provided written consent prior to participating in the study and were informed about their right to withdraw at any time without consequences. This study received ethical approval, using the ethical self-check from the Free University Amsterdam Ethics, ensuring compliance with ethical guidelines and standards. 

\subsection{Experimental Design}
For this study, a within-subject design was used to assess the user experience. The experiment consists of two conditions: a group of participants with a personality-driven robot, the personality condition, and a group with a task-oriented robot, the control condition. \\

In the first condition, the personality condition, the robot has its own endeavor, meaning it incorporates jokes during the interaction with the user. The robot was programmed to ask the user questions from a pre-provided medical questionnaire (see \ref{Appendix A: Participant instruction}), thus creating a personal interaction with the user. Throughout this interaction, the robot replied to the answers given by the user with witty remarks related to the users' response.\\

The second condition, the control condition, worked with a robot that is solely task-oriented and does not have an integrated personality. During the experiment, it asked the user the medical questions sequentially without any additional commentary. The user answered each question, hereafter the robot moves on to the next question. The robot did not say or do anything else, establishing a baseline for the analysis. \\

As this experiment consisted of only twelve participants for the experiment testing, each participant answered the questions in both conditions. To control for potential biases, participants were randomly assigned to one of the two groups. The first group interacted with the robot without a personality first, and the other half interacted with the robot with a personality first. Prior to the experiment, each participant was provided with written instructions on what they should do. As a result, each candidate was given an identical set of instructions to ensure the experiment's soundness. At the end, each participant filled out a questionnaire about their experience. 

\subsection{Materials}
The experiment setting was situated in a quiet room in the \textit{Nieuwe Universiteitsgebouw} at the \textit{Vrije Universiteit Amsterdam}. The participant was situated next to the robot, who was standing on the floor. Next to the participants was a table with a laptop that was used to capture the audio of the participants' responses. We use one robot for both condition in the experiment. A laptop was connected to the robot on which the code was run. The personality of the robot, which changed per condition, was controlled within the code.\\

A set of instructions was prepared which each participant received before their interaction with the robot. These instructions were very precise and clear to avoid confusion as that may introduce inconsistencies into the experiment. Additionally, these instructions included a disclaimer that indicated that all gathered data would be kept anonymous. All participants were unaware of the intentions of the robot in both conditions, as this hopefully results in acquiring the most accurate results from the final questionnaire. The full instructions can be found in appendix \ref{Appendix A: Participant instruction}.\\

\subsubsection{General health checkup questionnaire}
During both conditions, the NAO robot asked a series of questions to the participant during the experiment. The question originated from a questionnaire regarding a general health check-up. These questions ranged from basic information such as 'How old are you?' to health-related inquiries like 'How many days a week do you exercise or play sports?' The full questionnaire can be found in Appendix ~\ref{Appendix B: Health questionnaire}.

\subsubsection{User experience questionnaire}
The user experience questionnaire, a comprehensive questionnaire divided into four parts, was used to evaluate participants' experiences in each condition. The questionnaire consisted of both quantitative and qualitative measurements. The section with the quantitative measurement consisted of a combination of the Positive and Negative Affect Scale (PANAS) questionnaire \cite{Watson1988} and the Godspeed questionnaire \cite{Bartneck2009}. Additionally, the qualitative measurements included general open-ended questions regarding the user's perception of the robot. \\

The survey was initiated with open-ended questions with regard to the perception of the robot. Next, an adjusted version of the PANAS questionnaire \cite{Watson1988} was used to indicate the participants's emotional state. The Godspeed questionnaire \cite{Bartneck2009} was employed to determine the user perception of the robot through several adjectives. These questionnaires were filled out twice, once per condition, by each participant. Afterwards, the participant filled out the final part, which included a manipulation check, followed by questions regarding the preference between the robots. The questionnaire ended with the opportunity for the participant to write any additional notes. The complete questionnaire is included in the appendix~\ref{Appendix C: User Experience questionnaire}.

\subsubsection{Code}
\begin{figure}
    \centering
    \includegraphics[width=1\linewidth]{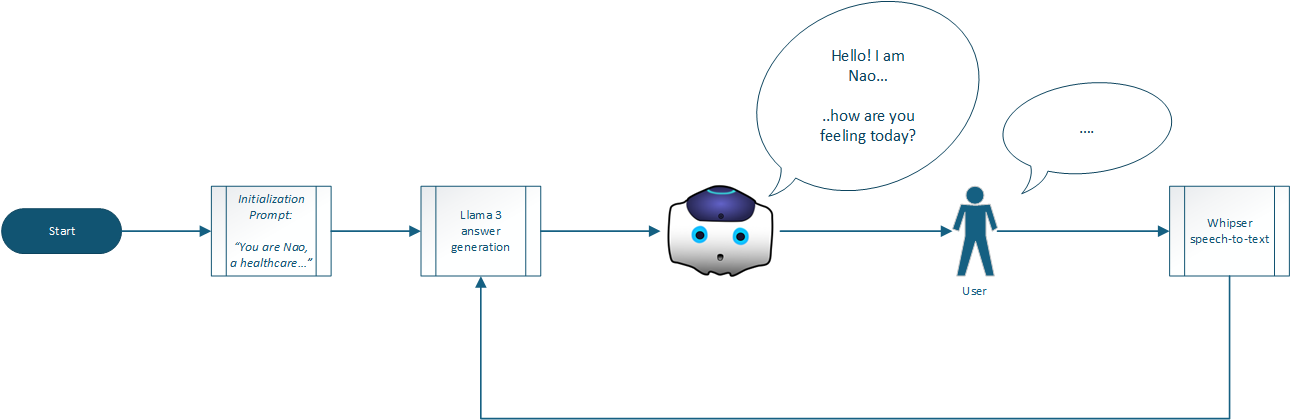}
    \caption{Conversational flow with NAO robot}
    \label{fig:conversation flow}
\end{figure}

In order to run the experiments, the NAO robot was programmed using the Social Interaction Cloud (SIC) \cite{SIC}, in combination with a local version of Meta's Llama 3 model \cite{Llama2024}, and OpenAI's whisper speech-to-text model \cite{whisper}. The general flow can be found in figure \ref{fig:conversation flow}.\\

The functionality of NAO is divided into two principal components. A version of Meta's Llama 3 model is employed to process the input of the initial prompt, as well as the user input. Llama is responsible for controlling the 'character' of NAO, and generating a new output to be pronounced by NAO based on the system prompt at the start of the interaction, or the user's feedback during the interaction. The text-to-speech module of the SIC framework is used to make NAO speak. \\

OpenAI's whisper \cite{whisper} is responsible for recording the user input and performing speech-to-text conversion, the output of which is then fed back to the Llama model. The interaction between the user and NAO can be summarized as follows:
\begin{enumerate}
    \item The Llama model is initialized with the system prompt
    \item The output of Llama is pronounced by NAO
    \item microphone input is recorded until the audio level drops below a certain threshold for a specified time, automatically ending the user's input cycle.
    \item Whisper is used to encode the audio into text
    \item Whisper's output is fed back into Llama
    \item Cycle repeats from point 2.
\end{enumerate}
During the design phase, several challenges were encountered. Initial testing showed that, though not unacceptable, it took a noticeable amount of time to generate a response after collecting user input, leaving the user waiting and unsure of progress. Previous work has already shown that in ordinary human-human interactions, a delay of more than 700 ms is perceived as long \cite{HumanTiming}. However, work From D. Kang et al. has indicated that the perceived waiting time can be reduced by providing some form of robot feedback \cite{RobotDelay}. To this extent, the NAO robot was programmed to indicate when it is listening for user input by switching its eye color from the default white to a warm pink tone. In an additional effort to limit the response time, the NAO robot was programmed with a local LLM, rather than a cloud-based alternative, to limit the response times of these services. In addition to response times, an effort had to be made to create a naturally flowing conversation. Rudimentary implementations of human-robot conversations involve preset durations for user speech input. The downside of this approach is that it severely limits the immersive-ness and user experience and creates an overall rigid feel. To mitigate this, a simple approach was employed in which the amplitude of the audio input was constantly measured and checked against a certain threshold.
\begin{figure}
    \centering
    \includegraphics[width=0.8\linewidth]{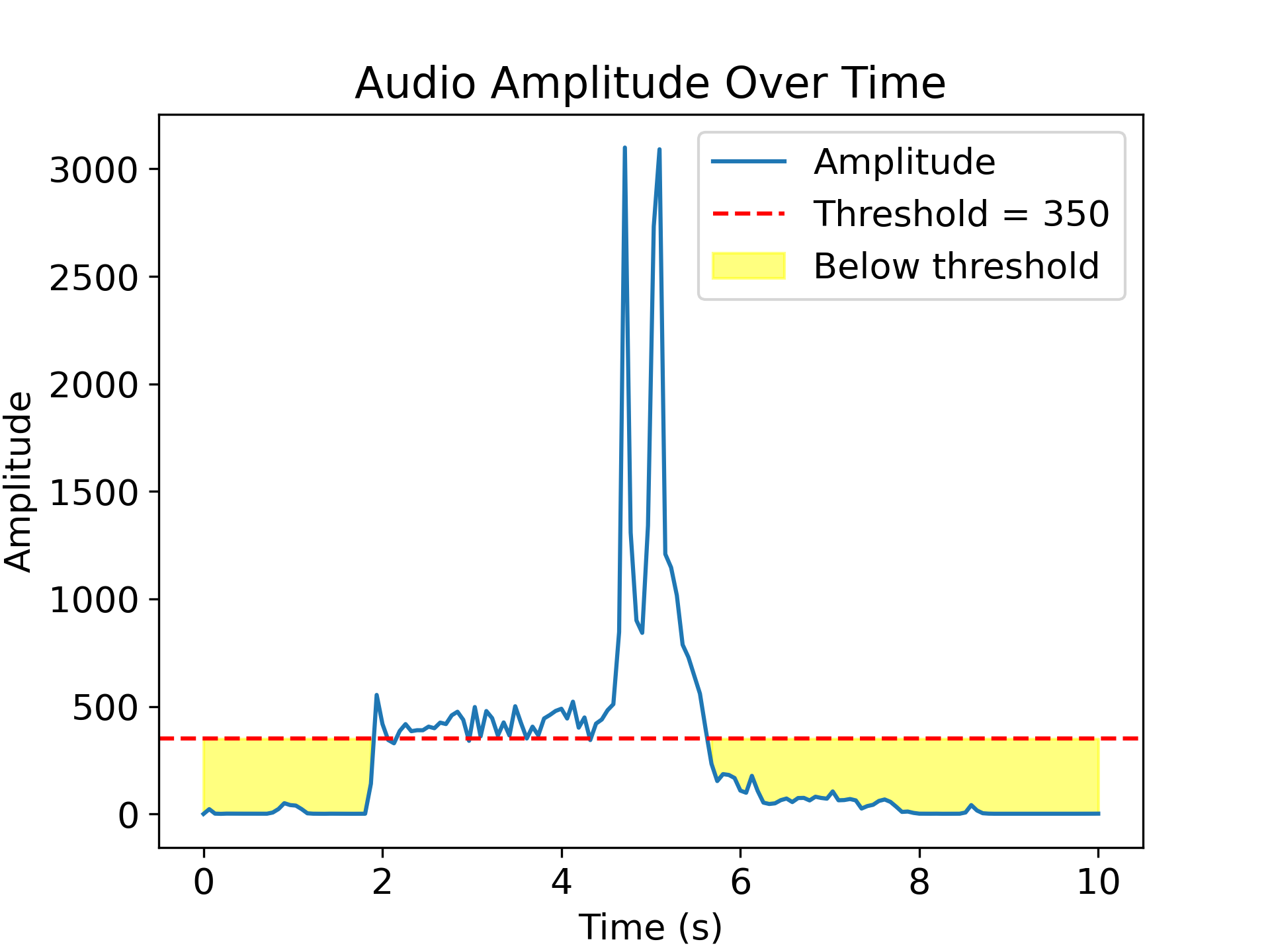}
    \caption{Amplitude of a sample response}
    \label{fig:audio-threshold}
\end{figure}
Figure \ref{fig:audio-threshold} shows the recording of a sample response. The yellow areas indicate occurrences during the response where the amplitude of the response was:
\begin{itemize}
    \item below a calibrated threshold (In this case 350Hz)
    \item For longer than a calibrated timespan (in this case 1.5 seconds)
\end{itemize}
  These parameters could be calibrated at any time in order to enhance a more natural flow of the conversation. If the above requirements were both satisfied, NAO automatically stopped recording and the audio was transcribed to text. However, during implementation, it was found that the NAO robot microphone was not of sufficient quality to enable this mechanism. Although it was possible to connect, the output format combined with an inherent background noise of NAO's own internal cooling equipment meant that instead the microphone of a laptop was used to capture the conversation audio.
\\
The open-source Llama model was selected in order to enhance the reproducibility of the overall experiment. The prompt used for the model can be found in appendix D.

\subsection{Procedure}
The experiment was initiated when the participant entered the room. They sat down on a chair facing the robot, who was standing on the floor. The participant received the instruction paper (see \ref{Appendix A: Participant instruction}), which they proceeded to read and sign. Next to the participants was a table with a laptop that was used to capture the audio of the participants' responses. When the code was initiated, the experiment started. The first group started with the personality condition, followed by the control condition. \\

In the personality condition, participants interacted with the robot which had its own endeavor. During this interaction, the robot asked medical questions and followed-up on each answer given by the user using some joke or witty remark, intended to relate to the user's response. Additionally, the prompt used to generate the exact phrasing of the questions was altered to encourage a more light-hearted and informal tone.\\

During the control condition, participants interacted with the robot that did not have its own endeavor. The robot asked them a question, waited for their answer, and continued with the following question. The prompt used to generate the exact phrasing of the questions was tuned to induce a formal and systemic tone.\\

After the first six participants, the second group started. They began in the control condition, followed by the personality condition. Hereafter, the participants were asked to move into the hallway, where they filled out our user experience questionnaire using Google Forms. 

\subsection{Data processing}
As the current study is a pilot study, we decided to not solely have quantitative measurement, but additionally include qualitative measurement. The questionnaires of the quantitative measurements were analyzed with a focus on descriptive statistics. As our sample is too small to make meaningful inferences with statistical tests, we opted for analyzing means and graphing them accordingly. The qualitative measures would give us a better understanding of the initial interaction with the robots, and most importantly, the preferences between the two conditions. The given answers were individually assessed and any interesting or particular outcomes were highlighted in the results. 

\section{Results}

\begin{table}[h!]
	\centering
	\caption{Descriptive Statistics PANAS}
	\label{descriptiveStatistics}
	{
		\begin{tabular}{lrrr}
			\toprule
			 &  & Mean & Std. Deviation  \\
			\cmidrule[0.4pt]{1-4}
			Interested & Personality & $4.083$ & $0.669$  \\
			 & Control & $2.750$ & $1.138$  \\
			Excited & Personality & $4.000$ & $0.739$  \\
			 & Control & $2.250$ & $1.055$  \\
			Upset & Personality & $1.417$ & $0.669$  \\
			 & Control & $1.333$ & $0.651$  \\
			Strong & Personality & $2.750$ & $0.754$  \\
			 & Control & $2.417$ & $1.165$  \\
			Enthusiastic & Personality & $4.000$ & $0.953$  \\
			 & Control & $2.250$ & $0.866$  \\
			Distressed & Personality & $1.917$ & $1.240$  \\
			 & Control & $1.583$ & $0.900$  \\
			Determined & Personality & $3.167$ & $0.937$  \\
			 & Control & $2.000$ & $1.044$  \\
			Nervous & Personality & $1.667$ & $1.155$  \\
			 & Control & $1.333$ & $0.888$  \\
			Alert & Personality & $1.917$ & $1.165$  \\
			 & Control & $2.000$ & $1.279$  \\
			Inspired & Personality & $3.083$ & $1.084$  \\
			 & Control & $1.667$ & $0.888$  \\
			\bottomrule
		\end{tabular}
        \label{ttt}
	}
\end{table}

\begin{figure}[h!]
    \centering
    \begin{minipage}[t]{0.48\textwidth}
        \centering
        \includegraphics[width=\textwidth]{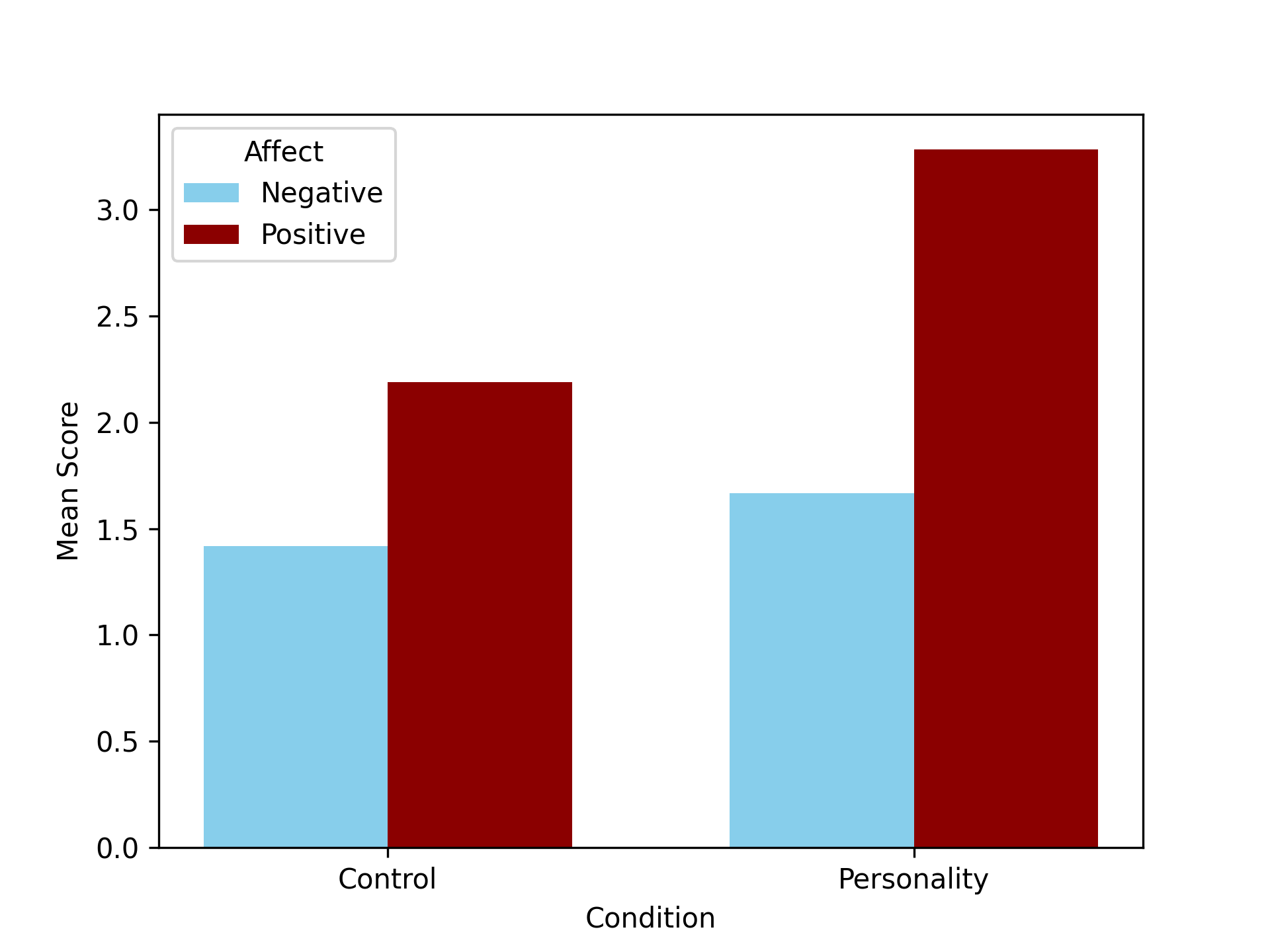}
        \caption{Graphical Representation of PANAS Scores}
        \label{Graph_PANAS1}
    \end{minipage}%
    \hfill
    \begin{minipage}[t]{0.48\textwidth}
        \centering
        \includegraphics[width=\textwidth]{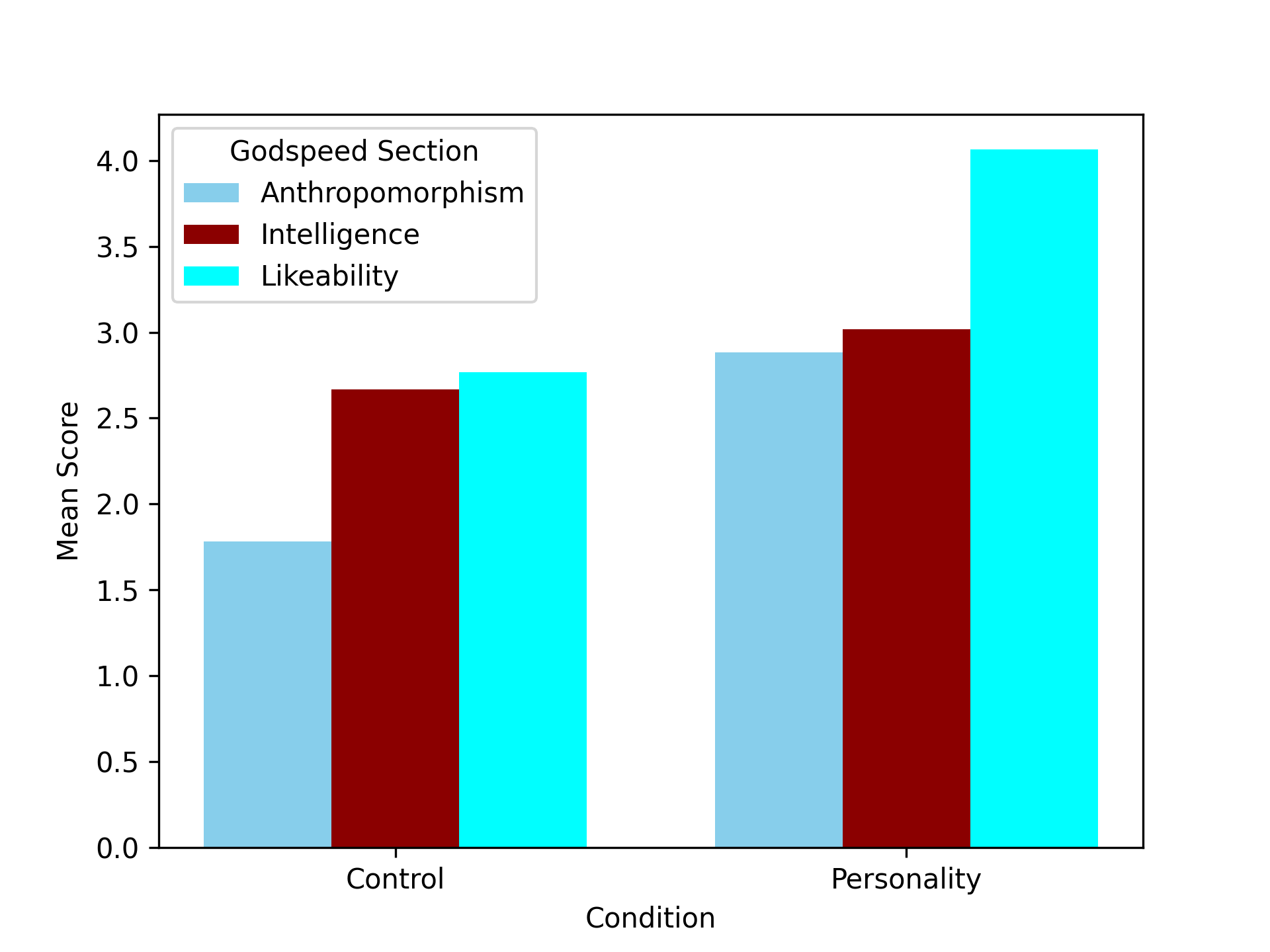}
        \caption{Graphical Representation of Godspeed Scores}
        \label{Graph_PANAS2}
    \end{minipage}
\end{figure}

\subsection{\textbf{Analysis of quantitative user feedback}}

The results suggest a clear difference between the control condition and the personality condition. Figure (\ref{Graph_PANAS1}) shows that the mean score for the negative affect is similarly quite low for both conditions. The mean positive effect, however, is clearly higher in the personality condition. This shows that the personality condition had a positive effect on the majority of the participants.
\\\\
Figure (\ref{Graph_PANAS2}) shows the mean anthropomorphism, intelligence and likability scores of both conditions. Participants rated these values from one to five, where one is not likable and five is very likable, for example. It can be seen that for the personality condition, the mean of each of these scores is higher than for the control condition. This suggests that the personality condition increased the user experience. Table (\ref{ttt}) gives a full overview of the results of our questionnaire.

\subsection{\textbf{Analysis of qualitative user feedback}}

All participants successfully detected the personality differences between the two robots. 

\subsubsection*{Personality condition}
Participants described the personality condition as more playful and conversational using such phrases as "more like a conversation" and "more playful". Several participants described the personality condition as "funny" and "enjoyable". This contributed to the interaction being perceived as generally more relaxed and engaging, and participants reported a preference for the personality condition by a large margin. However, this playfulness and sense of humor came with drawbacks. One participant describes the personality condition as "a little awkward personality-wise". Additionally, cultural differences played a negative role in the perception of the personality condition's humor with one participant experiencing a negative reaction to being called a 'potato'. Additionally, some participants felt like Robot's jokes were inappropriate for a medical setting.

\subsubsection*{Control condition}
The control condition was perceived as more professional and direct. The interaction with the control condition was frequently likened to filling out a "form" or engaging with a "humanized form". However, the control condition's formal tone contributed to participant disengagement, with participants describing the robot as "static", "boring", "not interesting" and "not engaging". This points to the tension between the objectives of preserving user engagement without undermining the practical goals of the interaction.

\subsubsection*{Comparison between the conditions}
When comparing the two robots, participants generally expressed a preference for the personality condition.
Participants described the experience as more relaxed and engaging and reported feeling more comfortable interacting with the personality condition. Meanwhile, the control condition's concise questions and formal tone caused the participants to adjust their response tone and make their responses more concise. This highlights how the robot's personality can have a significant impact on the interaction by causing a change in user behavior.

\section{Discussion}
This research concerns a humorous NAO robot that functions as a healthcare assistant, with Llama running as its functioning model for conversational interaction. Our results mentioned in the sections above state that NAO's humorous personality created a bonding user experience, it created an environment that encouraged participants to engage more openly and comfortably in the interaction. Humor seems to be an effective way to engage participants in the conversation, which is optimal for a healthcare environment where emotional support comes in useful.
\\\\
The participants mostly responded positively to the humor of NAO, making the interaction feel less clinical and less formal overall. This is in accordance with previous research findings, which stress how important humor is for interaction between humans \cite{Meyer} but, therefore, certainly between humans and robots. However, humor does not always provide benefits. In some cases, participants felt the humor did not fit the context and also because humor varies from person to person, it was more difficult for some to understand. This highlights the importance for NAO to properly tune out contextually what works and what does not, especially, for a healthcare robot.
\\\\
The choice of the Llama model ultimately worked out well. The model, though limited in size, proved to be accurate enough for this study. The ability to incorporate a touch of humor in between the serious answers had exactly the effect that matched the hypothesis. Nevertheless, the model used in this study was too small to deal with ambiguity in participant answers. This occasionally caused the model to produce answers which had nothing to do with the participants answer.
\\\\
Aside from the accuracy of the model, the complexity of the prompts in combination with limited computing resources induced a delay between processing the user response and subsequently generating and pronouncing a witty response, sometimes creating confusion or prolonged awkward silences. Future improvements to the interaction design should include clearer indications of activity such as animated eyes, a reduced processing time by using more advanced models and hardware, or a combination of the two. It was, however, observed that the mechanism for indicating when to answer (e.g. the switching of eye colors to pink) worked well in managing user expectations and orchestrating a smooth flow during the conversation.

\subsection{Limitations}
While the results look promising, for this study several limitations should be acknowledged.

\begin{itemize}
    \item \textbf{Sample Size and Diversity}: The study's sample consisted only of twelve university students, which may limit the generalization of the findings. Besides that, the participating students are in the same class, so they have already worked with these robots. This may lead to biases. In the future, it should also be tested on broader populations like older persons, diverse cultural groups, and people who already work in healthcare.
    \item \textbf{Interaction Duration}: The relatively short interaction time may not capture long-term user experiences. Longer interactions are necessary, as this gives a better idea of to what extent humor works with patients to put them at ease.
    \item \textbf{Model Training}: Even though the Llama model used was promptly tested to be a healthcare robot, in the future it is wise to fine-tune the model itself so that the LLM understands medical terms, ideologues, and patient sensitivity.
    \item \textbf{Robot Capabilities}: The robot's limited ability to understand complex user inputs may have affected user perceptions by occasionally misinterpreting a user's response and therefore producing non-nonsensical outputs.
\end{itemize}

\section{Conclusion}
This research paper set out to answer the research question: \textit{To what extent does adding personality to a questionnaire robot increase user experience?} A pilot study was proposed and executed making use of an NAO robot running an LLM with the two different conditions. Challenges were encountered during the design phase of the NAO robot including controlling the duration of user input with background noise, providing user feedback during processing on NAO's side, and overall response time. Results from the experiments indicate that not only were all users able to identify a clear difference in the behavior of NAO during both conditions, but users overall did feel more comfortable when interacting with the robot during the 'personality condition' and felt more engaged and open during question answering, thus confirming our original hypothesis. Limitations with regards to the experimental setup were discussed, and improvements to the overall design were suggested.

\section{Future Research}

As Foundation models move towards multi-modality, new capabilities are bound to emerge that may push forward the field of socially interactive robotics. The most recent multi-modal, speech-to-speech models built by OpenAI, for instance, display never-seen-before levels of dialogue management proficiency, interruption handling, turn-taking, and overall parsimony. We predict extending modalities further such as the addition of vision would enable breakthroughs in gaze tracking, face detection, emotion detection, and gesture recognition, further enhancing the robot's affective capabilities. 
\\\\
Additionally, further research needs to be conducted to test robot personality's impact on the user experience of diverse groups of participants from various cultures, age groups, and social backgrounds to help further contextualize robot behavior and adapt it to our needs. Additionally, the Llama model should be fine-tuned on healthcare-appropriate interactions (see limitations, Model Training section). One step further would be to use larger and more powerful models such as those from OpenAI, which may lead to superior, more "human-like" results.

\newpage
\appendix
\section{Appendix: Participant instruction}\label{Appendix A: Participant instruction}

\noindent
\textbf{Title of the Study:} Existential Crisis: A Social Robot’s Reason for Being \\

\noindent
\textbf{Research Team:} \\
Dora Medgyesy, Joella Galas, Julian van Pol, Rustam Eynaliyev and Thijs Vollebregt \\
Vrije Universiteit Amsterdam\\

\noindent
Thank you for participating in our study. Below you can find the instructions for this experiment and additional information regarding the informed consent.\\\\
\textbf{Study Procedures}
\begin{itemize}
    \item You will interact with two versions of our NAO robot.
    \item Both versions will ask you a series of medical questions.
    \item You start with the first version, then after 2 minutes, there will be a small break. Afterwards, you will have another interaction with the second version of the robot.
    \item After your interaction with the robot, you will complete a questionnaire to provide feedback on your experience.
\end{itemize}
\textbf{Duration}
The entire session will take approximately \textbf{10 minutes}, including the interaction with the robot and the completion of the questionnaire afterwards.\\\\
\textbf{Informed Consent}\\
Your participation in this study is entirely voluntary. You have the right to withdraw at any time without any consequences. Please read the following statements carefully and indicate your consent by signing below:
\begin{enumerate}
    \item I understand that my participation is voluntary and that I can withdraw at any time without penalty.
    \item I understand that my responses will be anonymous and that no personal data will be collected that could identify me.
    \item I understand that the data collected will be used solely for research purposes and may be published in academic journals.
    \item I have been informed about the nature of the study and the procedures involved.
\end{enumerate}

\noindent
\textbf{Consent} \\
By signing below, you indicate that you have read and understood the information provided and agree to participate in this study.

\noindent
\begin{tabbing}
\hspace{4cm} \= \hspace{5cm} \= \kill
\textbf{Participant Name:} \> \rule{5cm}{0.4pt} \\
\textbf{Participant Signature:} \> \rule{5cm}{0.4pt} \\
\textbf{Date:} \> \rule{5cm}{0.4pt}
\end{tabbing}

\vspace{2em}

\noindent
Thank you for your participation!

\section{Appendix: Health questionnaire}\label{Appendix B: Health questionnaire}
\begin{itemize}
        \item What is your name?
        \item How old are you?
        \item What was your bachelor's study?
        \item Do you have any allergies?
        \item Do you have High or Low Blood Pressure?
        \item Have you ever broken a bone?
        \item Do you wash your hands before eating and after using the restroom?
        \item Have you visited a doctor in the last 6 months?
        \item How many alcoholic beverages do you consume in a week?
        \item How many days a week do you exercise or play sports?
    \end{itemize}

\section{Appendix: User Experience questionnaire}\label{Appendix C: User Experience questionnaire}

\subsection*{Part 1: Describe the Robot – User Experience}

\textbf{Instructions}: Please answer the following questions based on your interaction with the robot.

\begin{enumerate}
    \item In your own words, describe your overall experience with the robot.
    \item What stood out to you the most during your interaction with the robot?
    \item How did the robot's behavior affect your engagement during the interaction?
    \item Did you feel the robot understood you? Please explain.
\end{enumerate}

\subsection*{Part 2: Numerical Questions – User Experience and Perception}

\textbf{A. How Do You Feel}

\textbf{Instructions}: Indicate to what extent you feel each emotion right now, after interacting with the robot.

Use a scale from 1 (\textit{Very slightly or not at all}) to 5 (\textit{Extremely}).

\begin{enumerate}
    \item Interested \dotfill \underline{\hspace{2cm}} (1--5)
    \item Excited \dotfill \underline{\hspace{2cm}} (1--5)
    \item Upset \dotfill \underline{\hspace{2cm}} (1--5)
    \item Strong \dotfill \underline{\hspace{2cm}} (1--5)
    \item Enthusiastic \dotfill \underline{\hspace{2cm}} (1--5)
    \item Distressed \dotfill \underline{\hspace{2cm}} (1--5)
    \item Determined \dotfill \underline{\hspace{2cm}} (1--5)
    \item Nervous \dotfill \underline{\hspace{2cm}} (1--5)
    \item Alert \dotfill \underline{\hspace{2cm}} (1--5)
    \item Inspired \dotfill \underline{\hspace{2cm}} (1--5)
\end{enumerate}

\textbf{B. How Do You View the Robot}

\textbf{Instructions}: For each pair of adjectives, select a number from 1 to 5 that best describes your perception of the robot.

\begin{enumerate}
    \item \textbf{Anthropomorphism}
    \begin{itemize}
        \item Fake \dotfill 1 --- 2 --- 3 --- 4 --- 5 \dotfill Natural
        \item Machinelike \dotfill 1 --- 2 --- 3 --- 4 --- 5 \dotfill Humanlike
        \item Unconscious \dotfill 1 --- 2 --- 3 --- 4 --- 5 \dotfill Conscious
        \item Artificial \dotfill 1 --- 2 --- 3 --- 4 --- 5 \dotfill Lifelike
        \item Moving rigidly \dotfill 1 --- 2 --- 3 --- 4 --- 5 \dotfill Moving elegantly
    \end{itemize}
    \item \textbf{Likeability}
    \begin{itemize}
        \item Dislike \dotfill 1 --- 2 --- 3 --- 4 --- 5 \dotfill Like
        \item Unfriendly \dotfill 1 --- 2 --- 3 --- 4 --- 5 \dotfill Friendly
        \item Unkind \dotfill 1 --- 2 --- 3 --- 4 --- 5 \dotfill Kind
        \item Unpleasant \dotfill 1 --- 2 --- 3 --- 4 --- 5 \dotfill Pleasant
        \item Awful \dotfill 1 --- 2 --- 3 --- 4 --- 5 \dotfill Nice
    \end{itemize}
    \item \textbf{Intelligence}
    \begin{itemize}
        \item Incompetent \dotfill 1 --- 2 --- 3 --- 4 --- 5 \dotfill Competent
        \item Ignorant \dotfill 1 --- 2 --- 3 --- 4 --- 5 \dotfill Knowledgeable
        \item Irresponsible \dotfill 1 --- 2 --- 3 --- 4 --- 5 \dotfill Responsible
        \item Unintelligent \dotfill 1 --- 2 --- 3 --- 4 --- 5 \dotfill Intelligent
        \item Foolish \dotfill 1 --- 2 --- 3 --- 4 --- 5 \dotfill Sensible
    \end{itemize}
\end{enumerate}

\subsection*{Part 3: Preference Between Robots}

\textbf{Instructions}: Please indicate your preference and provide reasons.

\begin{enumerate}
    \item Did you perceive Robot A as more task-oriented and Robot B as more social and engaging?
    \begin{itemize}
        \item [$\Box$] Yes
        \item [$\Box$] No
        \item [$\Box$] Unsure
    \end{itemize}
    \item Which robot did you prefer interacting with?
    \begin{itemize}
        \item [$\Box$] Robot A: Task-oriented
        \item [$\Box$] Robot B: Social and engaging
    \end{itemize}
    \item Please explain why you preferred this robot.
    \item Would you recommend your preferred robot to others? \dotfill \underline{\hspace{2cm}} (1--5)
    \begin{itemize}
        \item Scale from 1 (\textit{Definitely not}) to 5 (\textit{Definitely yes})
    \end{itemize}
\end{enumerate}

\subsection*{Part 4: Additional Notes}

Please provide any additional comments or suggestions about your experiences with the robots.

\bigskip

[Space for open-ended response]
\section{LLM prompts}
\subsubsection{initialization prompt - personality}
\textit{You are Nao, a healthcare assistant tasked with interviewing new patients, please introduce yourself! your output will be spoken using text to speech, do not write sound effects. the questions to ask will be provided later do not ask any questions for now. focus on introducing yourself only.}
\subsubsection{question prompt - personality}
\textit{Formulate a single question to find out the following: "{question}" your reply should only consist of a single question do not generate multiple questions your tone is polite and pleasant your question is directly aimed at the user in the first person}
\subsubsection{initialization prompt - control condition}
\textit{You are Nao, a healthcare assistant tasked with interviewing new patients, please introduce yourself! your output will be spoken using text to speech, do not write sound effects. the questions to ask will be provided later do not ask any questions for now. focus on introducing yourself only. be as monotone as possible.}
\subsubsection{question prompt - control condition}
\textit{Formulate a single question to find out the following: "{question}" your reply should only consist of a single question do not generate multiple questions your tone should= be formal and direct your question is directly aimed at the user in the first person}
\section{Appendix:  Individual contribution summary}\label{Appendix D:  Individual contribution summary}

\textbf{Dora Medgyesy}
\\\\
At the beginning of the project I contributed to discussions about which project we should choose and what we should do for our research. I connected to NAO with my laptop and experimented with some of the tutorials given to us to get a better understanding of the possibilities that can be achieved. Once we decided to use a large language model for our experiment, I did some research about the possibilities. I experimented with using open AI's models as well as using ollama. We decided to use ollama for our project. Then I worked on implementing Whisper. 
\\\\
I actively participated in group discussions about how we will do our research weekly. I helped decide how we will conduct our experiment and what our statistical analysis will be. One other group member and I were responsible for deciding the medical questionnaire that NAO will ask the participants. As a result I did research into what should be asked and what was appropriate. I also helped evaluate the ethical issues that may arise during our study and I contributed to the ethical self-check. 
\\\\
During our experiment another group member and I were responsible for choosing participant and ensuring that each participant filled out our questionnaire. I made sure that everyone felt comfortable and I answered any arising questions. 
\\\\
I contributed to writing the paper and I created the poster. I also read our paper many times and gave feedback to sections other group member wrote. I helped to prepare for the presentation and actively participated. 
\\\\
\textbf{Joella Galas}
\\\\
A Throughout the entire project I took on an organising role within our group. I was making a planning and setting deadlines for our group, made task divisions for everyone and I made sure everyone knew what they had to do before the next session. During the tutorials, I always tried to evaluate what everyone did, how it went and whether they needed any assistence. \\

After all of us managed to make the SIC environment work, I was in a group who would research Whisper. I managed to made it work on my laptop, wrote a code for it and send it to Thijs, who would connect it with OpenAI. \\

While we were coding, I was brainstorming with the group on what our research question was supposed to be. I came up with the research setting, the NOA robot asking questions for a questionnaire in a medical environment, and after some back and forth with the supervisor. I, with one of my groupmates, decided on the final research question and hypothesis. \\

When looking into the measurements, I was looking for possible questionnaire for the user experience questionnaire. I looked into possible psychology related questionnaires like the PANAS. Afterwards, I went over the possibilities with a groupmate. After he put together a full questionnaire, I checked and refined it. I put it into a forms for the experiment. Hereafter, when getting everything ready for the experiment I also wrote the participant instructions. I checked with the others whether we were all ready to start the experiment the next session.\\

Regarding the writing of the paper, I started out by writing my part of the methodology and checking and giving feedback to the introduction other people wrote. After the experiment, I refined what was needed, created the graphs for the analysis and gave feedback to the entire document. I edited some texts where I saw it was necessary and helped whenever someone needed help.
\\\\
\textbf{Julian van Pol}
\\\\
I spent the first week of the project getting NAO up and running and becoming familiar with its operations. I also participated in discussions about what interaction problem we were going to solve as a group and, thus, what experiment we were going to perform that we had to write a paper about.
\\
After everything worked, we decided to use Meta's Llama instead of OpenAI's well-known LLM for our interaction problem. Part of the group got to work with Llama and I and another groupmate looked at how to get Whisper working since the participants' output was going to be used as input for the LLM. In addition, it is also important to determine what questions NAO would ask the participants to test whether the robot affects the participants. I along with a team member again was going to look at the questionnaire that the robot was going to ask, in the end, we decided together which questions it was going to be. We did this based on the ethical self-check we were given for the report that our experiment had to meet. So I myself also did the ethical self-check.
\\\\
Further, throughout the whole project, I participated in all discussions and was there in every working group to actively think along with the decisions we had to make as a team. In addition, I also wrote parts of the report. For example, I wrote the introduction with another teammate. I also wrote the discussion, limitations, and future work to conclude the report.
\\\\
In addition to my own sections for the report, I also carefully read through sections from my teammates and tried to provide feedback as best I could. I also improved their feedback back into my sections. Moreover, I  rewrote sections when sometimes things did not go well or when there were some minor grammatical mistakes.
\\
During the experiment, I sat in the room with the participant and paid attention to whether everything was going as it should.
\\\\
\textbf{Rustam Eynaliyev}
\\\\
Throughout the project, I contributed to defining the research question, hypothesis, methodology and writing of design paper. During the first weeks, I integrated SIR framework with Open AI's GPT models, which we later decided to abandon in favor of Meta's Llama models.
\\\\
I actively participated in team discussions and as project progressed, I focused on performing a review of relevant SRI literature. I researched industry standard robot perception studies (e.g., Godspeed, RoSaS, Almere model, NARS) and gathered user survey questions. 
\\\\
I have refined and written multiple aspects of our design document including abstract, introduction, related work, robot questionnaires and future directions sections. I have reviewed and provided feedback to the rest of the team and incorporated their feedback. During the experiment, I assisted with the conducting of experiment, interviewing of users and subsequently conducting qualitative analysis of the results of the study and writing the qualitative analysis section.
\\\\
\textbf{Thijs Vollebregt}
\\\\
I spent the first two weeks of the course settings up the python environment and trying different methods for interacting with NAO (e.g. via the eyes, movement, animations etc.). Afterwords we were assigned the interaction problem and I took care of being the vocal point in communications with the topic supervisor, Jorrit Thijn. During the initial design phase we were debating on wether to use openAI's ChatGPT or a locally run LLM such as Llama. During this phase I invested time in geting a Llama model running, and selecting an appropriate model where I had to balance response time and model accuracy. After this step was complete I invested most of my time implementing the main script. I developed the overall flow, and logic that was used to identify whenever a user was done talking (as in figure 2). After the several separate compoments were ready (e.g. the answer generation from Llama, the whisper component for speech to text, and the user input recording) I combined all into a single master script, and dedicated the rest of my time of the implementation phase to focus on prompt engineering. Finally I contributed on the document where I wrote down and explained the various aspects I had developed, and assisted on other topics such as the experimental setup.
\end{document}